%% file: main.tex
\documentclass[10pt,twocolumn,letterpaper]{article}

\usepackage[pagenumbers]{iccv} % To force page numbers, e.g. for an arXiv version


\definecolor{iccvblue}{rgb}{0.21,0.49,0.74}
\usepackage[pagebackref,breaklinks,colorlinks,allcolors=iccvblue]{hyperref}
\usepackage{booktabs, caption, multirow, array}
\usepackage{colortbl}
\usepackage{adjustbox}
\usepackage{makecell}

\def\paperID{3878} % *** Enter the Paper ID here
\def\confName{ICCV}
\def\confYear{2025}

\title{Hierarchical Cross-modal Prompt Learning for Vision-Language Models}

\author{
	Hao Zheng\textsuperscript{\rm 1}, 
	Shunzhi Yang\textsuperscript{\rm 2},
	Zhuoxin He\textsuperscript{\rm 1},
	Jinfeng Yang\textsuperscript{\rm 2}, 
	Zhenhua Huang\textsuperscript{\rm 1}\thanks{Corresponding author.}\\
	\textsuperscript{\rm 1} South China Normal University,
        \textsuperscript{\rm 2} Shenzhen Polytechnic University\\
	{\tt\small \{zzeo.zheng, yangshunzhi1994\}@gmail.com}\\
        {\tt\small \{hezhuoxin37, huangzhenhua\}@m.scnu.edu.cn, jfyang@szpu.edu.cn}
}

\begin{document}
\maketitle

\begin{abstract}
Pre-trained Vision-Language Models (VLMs) such as CLIP have shown excellent generalization abilities. However, adapting these large-scale models to downstream tasks while preserving their generalization capabilities remains challenging. Although prompt learning methods have shown promise, they suffer from two fundamental bottlenecks that limit generalization: (a) modality isolation, and (b) hierarchical semantic decay. To address these limitations, we propose HiCroPL, a \textbf{Hi}erarchical \textbf{Cro}ss-modal \textbf{P}rompt \textbf{L}earning framework that establishes bidirectional knowledge flow between text and vision modalities, enabling them to refine their semantics mutually. HiCroPL routes knowledge flows by leveraging the complementary strengths of text and vision. In early layers, text prompts inject relatively clear semantics into visual prompts through a hierarchical knowledge mapper, enhancing the representation of low-level visual semantics. In later layers, visual prompts encoding specific task-relevant objects flow back to refine text prompts, enabling deeper alignment. Crucially, our hierarchical knowledge mapper allows representations at multi-scales to be fused, ensuring that deeper representations retain transferable shallow semantics thereby enhancing generalization. We further introduce a lightweight layer-specific knowledge proxy to enable efficient cross-modal interactions. Extensive evaluations across four tasks demonstrate HiCroPL's superior performance, achieving state-of-the-art results on 11 benchmarks with significant improvements.
Code is available at: \href{https://github.com/zzeoZheng/HiCroPL}{https://github.com/zzeoZheng/HiCroPL}.
\end{abstract}

\section{Introduction}
\label{sec:intro}

The advent of Vision-Language Models (VLMs) like Contrastive Language-Image Pretraining (CLIP)~\cite{CLIP} has revolutionized visual representation learning~\cite{he2020momentum, bengio2013representation, yang2024learning}. By aligning web-scale image-text pairs through contrastive pre-training, these models achieve remarkable zero-shot generalization via handcrafted prompts (\eg, ``a photo of a [class]" in CLIP~\cite{CLIP}). However, fine-tuning VLMs for downstream tasks remains challenging due to their massive scale, particularly in limited supervision. Prompt engineering~\cite{sahoo2024systematic} offers a lightweight alternative, but it depends on domain-specific priors and struggles to capture task-specific nuances.

\begin{figure*}[t]
    \centering
    \includegraphics[width=\linewidth, trim=0 180 0 200,clip]{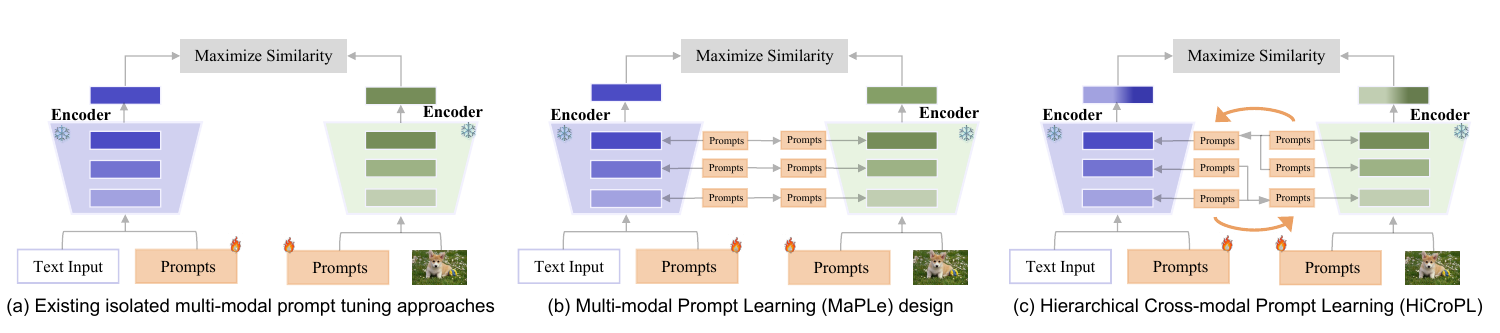}
    \vspace{-20pt}
    \caption{Comparison of HiCroPL with existing prompting approaches. (a) Most existing methods adopt uni-modal adaptation or isolated multi-modal solutions to fine-tune CLIP. (b) Multi-modal Prompt Learning (MaPLe) proposes a \textit{one way} (\ie text-to-vision) coupling function to bridge the two modalities, but visual concepts lack pathways to guide textual semantics. (c) HiCroPL introduces a bidirectional knowledge flow mechanism between the two modalities, enabling them to refine their semantics mutually for deep alignment. Besides, the representation used for downstream decisions contains rich intermediate features for improved generalization.}
    \label{fig:motivation}
    \vspace{-10pt}
\end{figure*}

This limitation has driven the evolution from static templates to learnable prompt paradigms.  CoOp~\cite{CoOp} pioneers this shift by optimizing context tokens, enabling task-specific adaptation through context optimization. Although effective in-domain, CoOp's design struggles with out-of-distribution generalization (\eg, new classes). CoCoOp~\cite{CoCoOp} mitigates this via image-conditioned prompts, dynamically adjusting to input visuals. Subsequent researches~\cite{KgCoOp, PromptSRC, ProGrad} further regularize prompt learning with frozen CLIP features. Despite progress, these methods share two fundamental bottlenecks that limit generalization:

(\textbf{a})\textbf{Modality Isolation}: Most methods adopt uni-modal adaptation~\cite{CoOp, CoCoOp, KgCoOp, jia2022visual} or isolated multi-modal solutions (Fig.~\ref{fig:motivation}(a))~\cite{PromptSRC, RPO} to fine-tune CLIP. Although MaPLe~\cite{MaPLe} proposes a \textit{one way} (\ie, text-to-vision) coupling function to bridge the two modalities, visual concepts lack pathways to guide textual semantics and remain isolated (Fig.~\ref{fig:motivation}(b)). This modality isolation hinders the mutual refinement of semantics between modalities, which is crucial for tasks requiring joint vision-language understanding~\cite{lin2023multimodality}.
%-------------------------------------------------------------------------

%-------------------------------------------------------------------------
(\textbf{b})\textbf{Hierarchical Semantic Decay}: Different levels in neural networks encode distinct types of knowledge and features~\cite{lin2017feature, bao2023lightweight}. For instance, shallow layers in VLMs capture task-agnostic low-level representations that exhibit strong cross-task transferability~\cite{MMA, neo2025towards}, while deep layers encode task-specific semantics~\cite{neo2025towards}. However, current approaches~\cite{PromptSRC, CoOp, wang2024learning, zhao2024learning} predominantly rely on final-layer features for downstream decisions, neglecting the rich hierarchical representations present in preceding layers. (Fig.~\ref{fig:motivation}(a) and (b)). This oversight stems from the lack of explicit mechanisms to synergize multi-scale features, leading to the decay of intermediate semantics during forward propagation and ultimately limiting generalization.

To address the dual challenges, we propose HiCroPL, a \textbf{Hi}erarchical \textbf{Cro}ss-modal \textbf{P}rompt \textbf{L}earning framework that establishes bidirectional knowledge flow between text and vision modalities, enabling them to refine their semantics mutually for deep alignment. HiCroPL routes knowledge flows by leveraging modality-specific strengths at different network depths. Specifically, in early layers, text prompts with relatively clear semantic information~\cite{MMA, li2021align} are mapped to visual prompts via a hierarchical knowledge mapper, enhancing low-level visual features. Conversely, in later layers, visual prompts encode task-specific semantics~\cite{neo2025towards, wang2024hpt++} and are mapped back to text prompts, grounding textual semantics in visual details for precise alignment. The entire pipeline of bidirectional knowledge flow constructs reciprocal pathways to facilitate the information exchange between text and vision modalities, enabling them to refine each other's semantics (\textbf{addressing the challenge (a)}). Simultaneously, the hierarchical knowledge mapper captures multi-scale semantic from cross-modal interactions, progressively integrating transferable representations from preceding layers to enhance generalization (\textbf{addressing the challenge (b)}). Finally, consistency regularization further preserves CLIP’s zero-shot capabilities, ensuring robust generalization.

Extensive evaluations across four tasks demonstrate HiCroPL’s superior performance. In the base-to-novel generalization task, HiCroPL outperforms the previous state-of-the-art method CoPrompt~\cite{CoPrompt} by 1.89\%, 0.76\%, and 1.28\% on the base classes, novel classes, and harmonic mean over 11 benchmark datasets, respectively. The key advantages of this paper include:
\begin{itemize}
  \item We propose a novel hierarchical prompt learning framework that effectively adapts VLMs to downstream tasks while preserving their inherent generalization capability.
  \item The bidirectional knowledge flow establishes reciprocal pathways between text and vision modalities, enabling mutual refinement of cross-modal semantics.
  \item The design of the hierarchical knowledge mapper facilitates information transfer between modalities at multiple scales, mitigates semantic decay, and improves generalization performance.
  \item Comprehensive experiments across 4 tasks and 11 benchmarks validate HiCroPL’s effectiveness and robustness. 
\end{itemize}
%-------------------------------------------------------------------------

\section{Related Work}
\label{sec:relawork}

\textbf{Vision-Language Models.}
Recent advances in nature language-supervised Vision-Language Models (VLMs) like CLIP~\cite{CLIP}, ALIGN~\cite{ALIGN}, and FLIP~\cite{FLIP} have established new paradigms in visual representation learning. Unlike traditional methods reliant on image-only supervision, these models learn joint visual-linguistic representations through self-supervised alignment of large-scale image-text pairs. Taking CLIP~\cite{CLIP} as an example, it consists of a text encoder and a vision encoder, each designed to encode features from its own modality. During pre-training, CLIP aligns approximately 400 million image-text pairs by minimizing a contrastive loss objective~\cite{oord2018representation}, which simultaneously pulls paired image-text embeddings closer in a shared multimodal space while repelling unpaired ones. While achieving remarkable zero-shot generalization, adapting VLMs to downstream tasks without compromising their innate capabilities remains an open challenge. Our approach exploits rich hierarchical semantics to maintain generalization performance.

\noindent\textbf{Prompt Learning for VLMs.} 
Initially proposed in NLP~\cite{liu2023pre, li2021prefix, lester2021power, sun2024benchmarking}, prompt learning has proven effective for adapting VLMs to downstream tasks~\cite{he2016deep, girshick2014rich, huang2022making, wanguni, lin2024multi}. By inserting learnable vectors into input or intermediate layers while keeping the backbone frozen, this technique mitigates catastrophic forgetting~\cite{robins1993catastrophic} and preserves zero-shot capabilities. The multi-modal nature of CLIP results in two types of prompt tuning strategies: text-based prompt tuning~\cite{ProGrad, CoCoOp, TCP, ProDA, CoOp} and multi-modal adaptation~\cite{MaPLe, PromptSRC, CoPrompt, RPO}. The former, pioneered by CoOp~\cite{CoOp}, optimizes learnable text prompts to provide task-specific context. CoCoOp~\cite{CoCoOp} generates image-conditioned prompts via a meta-network to address the weak generalization issue of CoOp~\cite{CoOp} on novel classes. KgCoOp~\cite{KgCoOp} and ProGrad~\cite{ProGrad} construct regularization terms in the text branch to constrain learnable prompts and general knowledge to avoid overfitting. TCP~\cite{TCP} proposes class-aware prompts to inject class-level knowledge into the prompts. The latter direction explores multi-modal adaptation, recognizing that isolated text tuning underutilizes CLIP’s cross-modal potential. MaPLe~\cite{MaPLe} proposes a multi-modal prompting framework to adapt both the vision and language branches of CLIP. RPO~\cite{RPO} introduces Read-only Prompt to prevent internal representation shift during adaptation. PromptSRC~\cite{PromptSRC} and CoPrompt~\cite{CoPrompt} employ additional loss functions to regularize the image and text branches separately.

We note that both types lack exploration of cross-modal collaboration, as they primarily tune the encoders independently (Fig.~\ref{fig:motivation}(a)). While MaPLe makes an effort, its one-way coupling function fails to fully exploit the interaction potential between modalities (Fig.~\ref{fig:motivation}(b)). In this work, we introduce a bidirectional knowledge flow mechanism to ensure the completeness of the cross-modal interaction, allowing the semantics of different modalities to mutually refine each other (Fig.~\ref{fig:motivation}(c)).

\section{Method}
\label{sec:method}
Following most existing works~\cite{CoOp, MaPLe, PromptSRC, CoPrompt, KgCoOp}, HiCroPL builds upon the pre-trained CLIP model~\cite{CLIP}, which utilizes transformer-based architectures for both the visual and text encoders. First, we introduce the preliminary knowledge of CLIP and prompt learning, followed by a detailed description of our proposed HiCroPL.
%-------------------------------------------------------------------------

\subsection{Preliminary}
The CLIP model, pre-trained on large-scale image-text pairs, has garnered significant attention from natural language processing and computer vision communities. It employs a dual-tower structure comprising a text encoder \textit{f}$_T$ and an image encoder \textit{f}$_I$. For open-vocabulary image classification, CLIP aligns visual and textual embeddings via cosine similarity, enabling zero-shot prediction. Formally, given a class \textit{c} from a dataset with \textit{N} classes, CLIP constructs a textual description using the pre-defined template s$_c$ = ``a photo of a [\textit{c}]". This is tokenized into discrete tokens t$_c$ = \textit{tokenizer} (s$_c$) and encoded as:  W$_c$ = \textit{f}$_T$(t$_c$) $\in \mathbb{R}^{\textit{d}_t}$, where ${d}_t$ represents the text feature dimension and W$_c$ corresponds to the [eos] token embedding serves as the class-specific text representation. On the visual side, an input image \textit{x} $\in \mathbb{R}^{\textit{H}\times\textit{W}\times3}$ is split into \textit{n} fixed-size patches and prepended with a class token. These patches and the class token are then projected into patch embeddings \textit{E}$ \in \mathbb{R}^{\textit{(n+1)} \times d_v}$. After processing by stacked transformer blocks, the final class token embedding V = \textit{f}$_I$(\textit{E}) $\in \mathbb{R}^{d_v}$ represents the global image semantics, where ${d}_v$ is the image feature dimension. The prediction probability is computed as follows:
\begin{equation}
\label{deqn_ex1a}
P(y=c|x) = \frac{\textrm{exp}(cos(V,W_{c}^{\mathbf{T}})/\tau )}{\sum_{n=1}^{N}\textrm{exp}(cos(V,W_{n}^\mathbf{T})/\tau)},
\end{equation}
where \textit{cos}($\cdot$) denotes the cosine similarity, $\tau$ is a temperature parameter, and $W_{c}$ represents the text embedding of the class $c$. 

Prompt learning~\cite{CoOp} adapts VLMs to downstream tasks by replacing handcrafted prompts with learnable vectors. In multi-modal prompt learning~\cite{MaPLe}, task-specific prompts are appended to both image and text inputs to align with CLIP’s architecture. On the text side, the static template ``a photo of a [class]" is replaced with a sequence of learnable tokens \textbf{$P_t$} = $\{p^1_t, p^2_t,...,p^m_t\}$, except for the class embedding, where \( p^i_t \in \mathbb{R}^{d_t} \) is a learnable text token and \( m \) denotes the number of learnable tokens. On the image side, the \textit{n} fixed-size patches are projected into embeddings $\{I_{cls}, I_1, I_2,...,I_n\}$ and concatenated with learnable visual prompts \textbf{$P_v$} = $\{p^1_v, p^2_v,...,p^m_v\}$, where \( p^i_v \in \mathbb{R}^{d_v} \) is a learnable image token, $I_{cls}$ and $ I_i$ are class token and patch embedding, respectively. The combined sequences $\{P_t, [class]\}$ and $\{P_v, I_{cls}, I_1, I_2,..., I_n\}$ are then fed to text and vision encoders, respectively, to extract prompted features. Recent studies~\cite{MaPLe, PromptSRC} have demonstrated the effectiveness of injecting learnable tokens into deeper layers. Specifically, for each layer \( l \in \{1, \dots, L\} \), a set of \textit{m} learnable tokens $\{p^{l,1}_t, p^{l,2}_t,...,p^{l,m}_t\}$ and $\{p^{l,1}_v, p^{l,2}_v,...,p^{l,m}_v\}$ are appended to the text and visual inputs, respectively. Here, $p^{l,i}_t$ denotes the \textit{i}-th token at layer \textit{l} for the text modality, while $p^{l,i}_v$ represents its visual counterpart.

%-------------------------------------------------------------------------

%-------------------------------------------------------------------------

\begin{figure*}[t]
    \centering
    \includegraphics[width=\linewidth, trim=35 125 48 120,clip]{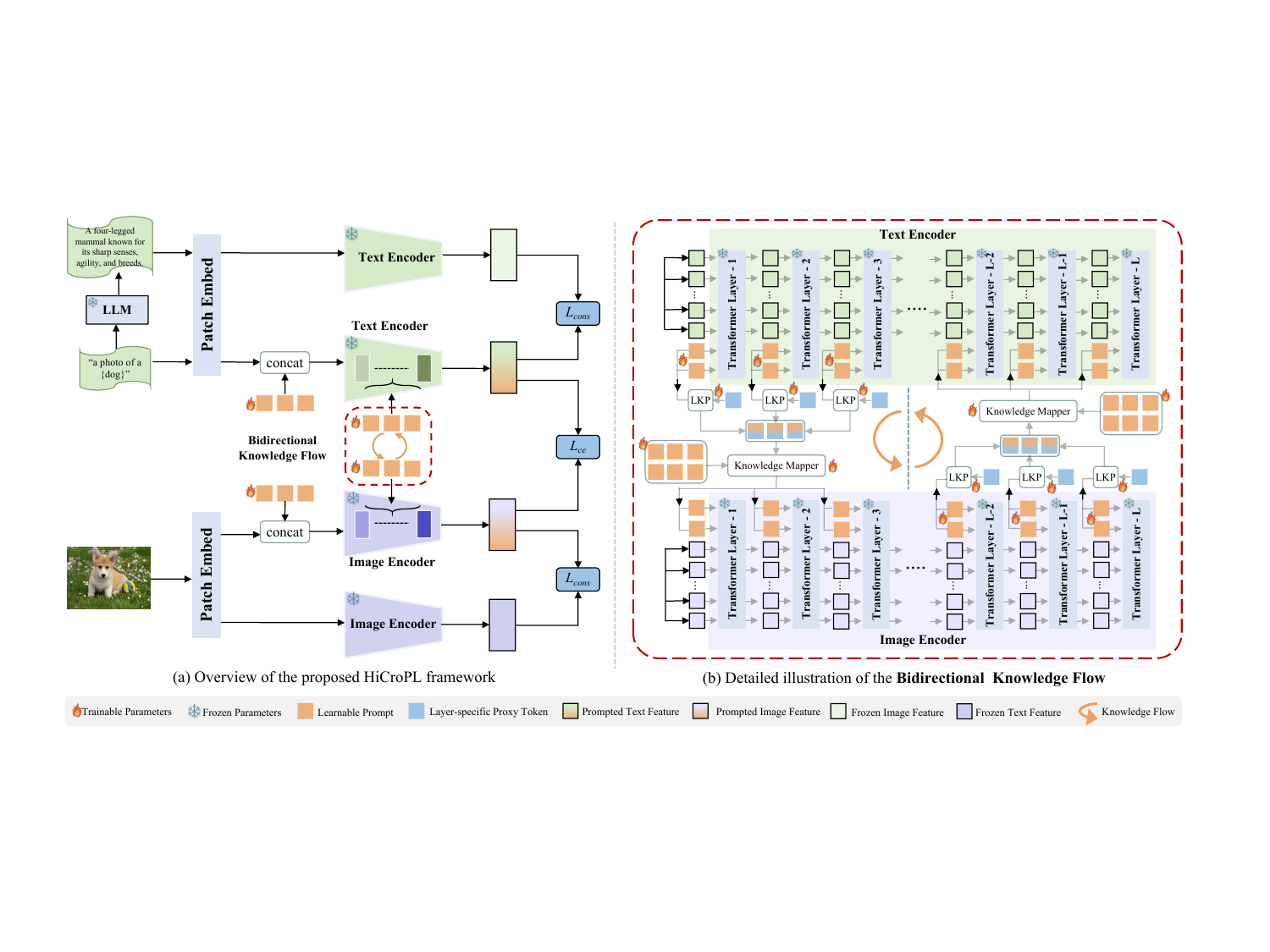}
    \vspace{-20pt}
    \caption{(a) Overview of the proposed HiCroPL framework. (b) Detailed illustration of the Bidirectional Knowledge flow mechanism. From Layer 1 to \textit{k}, the LKP first initializes layer-specific proxy tokens to encapsulate the key information relevant to the current layer, which then guide visual prompt refinement via the mapper $\mathcal{M}$. The reverse flow from Layer \textit{k}+1 to \textit{L} follows an identical process.}
    \label{fig:framework}
    \vspace{-10pt}
\end{figure*}

%-------------------------------------------------------------------------

\subsection{Hierarchical Cross-modal Prompt Learning }
Multi-modal prompting pushes prompt learning towards a solution of dual-encoder tuning to align with CLIP’s architecture. Subsequent works~\cite{PromptSRC, RPO, cho2023distribution} follow this paradigm, but they overlook a critical concept mentioned in MaPLe~\cite{MaPLe}: cross-modal synergy. This confines them to independently tuning text and visual encoders, limiting cross-modal information interaction. Furthermore, existing methods fail to integrate multi-scale semantics, relying solely on high-level task-specific features for downstream decisions. In reality, low-level features in intermediate layers encode rich, transferable representations~\cite{MMA, zhang2024concept, zhou2021domain, huang2022feature}, such as colors and shapes, which are crucial for generalization. To address these limitations, we propose HiCroPL, as illustrated in Fig.~\ref{fig:framework}(a). HiCroPL introduces a bidirectional knowledge flow mechanism (Fig.~\ref{fig:framework}(b)) that enables mutual refinement of text and visual prompts. Concurrently, the hierarchical knowledge mapper facilitates cross-modal feature mapping at multiple scales, ensuring that low-level features directly influence prompts. This allows the final-layer representation to contain rich information from the intermediate layers, mitigating their decay during forward propagation. We provide a comprehensive explanation of HiCroPL’s design in further detail below.

\noindent\textbf{Bidirectional Knowledge Flow.} 
Human perception relies on the synergistic interplay between linguistic and visual modalities~\cite{lin2023multimodality}. With visual concepts (\eg, a photo of a ``Boeing-737"), we can easily characterize an aircraft. Instead, it's effortlessly to identify a papillon dog among diverse canine images through descriptive textual prompts. Inspired by this reciprocity, HiCroPL establishes bidirectional knowledge flow to emulate this natural interplay, where knowledge flows are hierarchically governed by the complementary strengths of each modality:

\textit{Text-guided vision refinement.} Text embeddings, encoded with semantic category names, exhibit stronger priors in shallow layers~\cite{MMA}. Thus, from Layer 1 to \textit{k}, text prompts inject discriminative semantics into visual prompts via Hierarchical Knowledge Mapper (detailed in the next). This leverages CLIP’s linguistic priors to refine low-level visual features, reducing the modality gap.

\textit{Vision-grounded text alignment.} As visual features progressively encode task-specific patterns in deeper layers~\cite{MMA, neo2025towards}. From Layer \textit{k}+1 to \textit{L}, these visually enriched prompts reflux to text prompts, grounding textual semantics in task-relevant visual concepts for precise alignment.

Unlike MaPLe’s one-way coupling, HiCroPL establishes reciprocal pathways to achieve completeness in modality synergy, where text and vision mutually refine each other’s representations.

\noindent\textbf{Hierarchical Knowledge Mapper.} Existing methods predominantly rely on final-layer features for downstream decisions, and fail to exploit rich hierarchical representations embedded in intermediate layers. To harmonize them, we propose a hierarchical knowledge mapper $\mathcal{M}$, which acts as a bridge between the modalities while enabling the prompts to fuse multi-scale features from another modality. Additionally, a lightweight Layer-specific Knowledge Proxy (LKP) is introduced to aggregate intra-layer prompts, allowing a single proxy token to encapsulate the key information relevant to the current layer, thereby enabling efficient mapping process. 

Since the two mapping processes are identical except for the direction, we illustrate the overall workflow using the text-to-image mapping as an example. For each layer \textit{l}, LKP first initializes a layer-specific proxy token $p^l_p$. The \textit{m} text prompts $P_t = \{p^{l,1}_t, p^{l,2}_t,...,p^{l,m}_t\}$ are then compressed into this proxy prompt via a light cross-attention. This process can be formulated as:
\begin{equation}
\label{equation:HKP}
\tilde{p^l_p} = \textrm{CrossAttention}(p^l_p, P_t, P_t) \quad l \in 1,2,...,k,
\end{equation}
Here, $\tilde{p^l_p}$ denotes the refined proxy token aggregating information from all \textit{m} text prompts at layer \textit{l}. This reduces the input dimensionality to the subsequent mapper $\mathcal{M}$ from \textit{m} \textbf{$\cdot$} \textit{k} to \textit{k} while preserving layer-specific contextual information. The visual prompts $P_v = \{p^{l,1}_v, p^{l,2}_v,...,p^{l,m}_v\}$ subsequently interact with all refined proxy tokens
$\tilde{P_p} = \{\tilde{p^{1}_p},\tilde{p^{2}_p},...,\tilde{p^{k}_p}\}$ through the mapper $\mathcal{M}$:
\begin{equation}
\label{equation:MSKM}
P_v = \mathcal{M}(P_v, \tilde{P_p}, \tilde{P_p}),
\end{equation}
where $\mathcal{M}$ is implemented as a multi-head attention module~\cite{vaswani2017attention} with layer normalization and feed-forward networks. This hierarchical fusion enables each visual prompt to assimilate multi-scale cross-modal knowledge. Compared to layer-to-layer projection, our approach allows prompts to retain general patterns from preceding layers for enhanced generalization. The Appendix provides a detailed formulation of mapper $\mathcal{M}$.

\noindent\textbf{Training Objective.} We utilize the cross-entropy loss function as a supervised loss for image classification:
\begin{equation}
\label{ce_loss}
\mathcal{L}_{ce} = -\mathrm{log}\frac{\textrm{exp}(cos(V,W_{c}^{\mathbf{T}})/\tau )}{\sum_{n=1}^{N}\textrm{exp}(cos(V,W_{n}^\mathbf{T})/\tau)}.
\end{equation}

Inspired by~\cite{CoPrompt, PromptSRC, KgCoOp}, we further introduce a consistency regularization term $\mathcal{L}_{cons}$ to align the frozen and prompted embeddings, thereby enhancing generalization. Specifically, the frozen text embeddings are derived from detailed class descriptions generated by a large language model (LLM) and encoded by the pretrained CLIP text encoder, while the frozen image embeddings are obtained by processing the input image through the pretrained CLIP vision encoder. The consistency loss is defined as:
\begin{equation} 
\label{cons_loss}
\mathcal{L}_{cons} = 2 - \cos((V+V_p), V_p) - \cos((W+ W_p), W_p), 
\end{equation}
where $V$ and $W$ represent the frozen image and text embeddings, respectively, and $V_p$ and $W_p$ are their corresponding prompted embeddings.
Finally, the overall loss function used for training is:
\begin{equation}
\label{final_loss} 
\mathcal{L} = \mathcal{L}_{ce} + \lambda \mathcal{L}_{cons}, 
\end{equation}
where $\lambda$ is a hyperparameter controlling the weight of the consistency loss.

%-------------------------------------------------------------------------

\begin{table*}[t]
\renewcommand{\arraystretch}{0.95}
  \centering
  \scriptsize
  \resizebox{\linewidth}{!}{
  \begin{tabular}{l ccc|ccc|ccc|ccc}
    \toprule
    \multirow{2}{*}{Method} & 
    \multicolumn{3}{c}{(a) Average} & 
    \multicolumn{3}{c}{(b) ImageNet} & 
    \multicolumn{3}{c}{(c) Caltech101} & 
    \multicolumn{3}{c}{(d) OxfordPets} \\
    \cmidrule(lr){2-4} \cmidrule(lr){5-7} \cmidrule(lr){8-10} \cmidrule(lr){11-13}
    & Base & Novel & HM & Base & Novel & HM & Base & Novel & HM & Base & Novel & HM \\
    \midrule
    CLIP       & 69.34 & 74.22 & 71.70 & 72.43 & 68.14 & 70.22 & 96.84 & 94.00 & 95.40 & 91.17 & 97.26 & 94.12 \\
    CoOp       & 82.69 & 63.22 & 71.66 & 76.47 & 67.88 & 71.92 & 98.00 & 89.81 & 93.73 & 93.67 & 95.29 & 94.47 \\
    CoCoOp     & 80.47 & 71.69 & 75.83 & 75.98 & 70.43 & 73.10 & 97.96 & 93.81 & 95.84 & 95.20 & 97.69 & 96.43 \\
    KgCoOp     & 80.73 & 73.60 & 77.00 & 75.83 & 69.96 & 72.78 & 97.72 & 94.39 & 96.03 & 94.65 & 97.76 & 96.18 \\
    MaPLe      & 82.28 & 75.14 & 78.55 & 76.66 & 70.54 & 73.47 & 97.74 & 94.36 & 96.02 & 95.43 & 97.76 & 96.58 \\
    PromptSRC  & \underline{84.26} & 76.10 & 79.97 & 77.60 & 70.73 & 74.01 & 98.10 & 94.03 & 96.02 & 95.33 & 97.30 & 96.30 \\
    TCP        & 84.13 & 75.36 & 79.50 & 77.27 & 69.87 & 73.38 & 98.23 & 94.67 & 96.42 & 94.67 & 97.20 & 95.92 \\
    MMA        & 83.20 & 76.80 & 79.87 & 77.31 & 71.00 & 74.02 & 98.40 & 94.00 & 96.15 & 95.40 & \underline{98.07} & 96.72 \\
    CoPrompt   & 84.00 & \underline{77.23} & \underline{80.47} & \underline{77.67} & \underline{71.27} & \underline{74.33} & \underline{98.27} & \underline{94.90} & \underline{96.56} & \underline{95.67} & \textbf{98.10} & \underline{96.87} \\
    \rowcolor{gray!20}\textbf{HiCroPL} & \textbf{85.89} & \textbf{77.99} & \textbf{81.75} & \textbf{78.07} & \textbf{71.72} & \textbf{74.76} & \textbf{98.77} & \textbf{95.96} & \textbf{97.34} & \textbf{96.28} & 97.76 & \textbf{97.01} \\
    \midrule
    \midrule
    \multirow{2}{*}{Method} & 
    \multicolumn{3}{c}{(e) StanfordCars} & 
    \multicolumn{3}{c}{(f) Flowers102} & 
    \multicolumn{3}{c}{(g) Food101} & 
    \multicolumn{3}{c}{(h) FGVCAircraft} \\
    \cmidrule(lr){2-4} \cmidrule(lr){5-7} \cmidrule(lr){8-10} \cmidrule(lr){11-13}
    & Base & Novel & HM & Base & Novel & HM & Base & Novel & HM & Base & Novel & HM \\
    \midrule
    CLIP       & 63.37 & 74.89 & 68.65 & 72.08 & \textbf{77.80} & 74.83 & 90.10 & 91.22 & 90.66 & 27.19 & 36.29 & 31.09 \\
    CoOp       & 78.12 & 60.40 & 68.13 & 97.60 & 59.67 & 74.06 & 88.33 & 82.26 & 85.19 & 40.44 & 22.30 & 28.75 \\
    CoCoOp     & 70.49 & 73.59 & 72.01 & 94.87 & 71.75 & 81.71 & 90.70 & 91.29 & 90.99 & 33.41 & 23.71 & 27.74 \\
    KgCoOp     & 71.76 & \textbf{75.04} & 73.36 & 95.00 & 74.73 & 83.65 & 90.50 & 91.70 & 91.10 & 36.21 & 33.55 & 34.83 \\
    MaPLe      & 72.94 & 74.00 & 73.47 & 95.92 & 72.46 & 82.56 & 90.71 & \underline{92.05} & \underline{91.38} & 37.44 & 35.61 & 36.50 \\
    PromptSRC  & 78.27 & \underline{74.97} & 76.58 & \underline{98.07} & 76.50 & \textbf{85.95} & 90.67 & 91.53 & 91.10 & \underline{42.73} & 37.87 & \underline{40.15} \\
    TCP        & \underline{80.80} & 74.13 & \underline{77.32} & 97.73 & 75.57 & 85.23 & 90.57 & 91.37 & 90.97 & 41.97 & 34.43 & 37.83 \\
    MMA        & 78.50 & 73.10 & 75.70 & 97.77 & 75.93 & 85.48 & 90.13 & 91.30 & 90.71 & 40.57 & 36.33 & 38.33 \\
    CoPrompt   & 76.97 & 74.40 & 75.66 & 97.27 & \underline{76.60} & \underline{85.71} & 90.73 & \textbf{92.07} & \textbf{91.40} & 40.20 & \underline{39.33} & 39.76 \\
    \rowcolor{gray!20}\textbf{HiCroPL} & \textbf{81.51} & \textbf{75.04} & \textbf{78.14} & \textbf{98.29} & 75.46 & 85.38 & \textbf{90.96} & 91.67 & 91.31 & \textbf{48.38} & \textbf{41.75} & \textbf{44.82} \\
    \midrule
    \midrule
    \multirow{2}{*}{Method} & 
    \multicolumn{3}{c}{(i) SUN397} & 
    \multicolumn{3}{c}{(j) DTD} & 
    \multicolumn{3}{c}{(k) EuroSAT} & 
    \multicolumn{3}{c}{(l) UCF101} \\
    \cmidrule(lr){2-4} \cmidrule(lr){5-7} \cmidrule(lr){8-10} \cmidrule(lr){11-13}
    & Base & Novel & HM & Base & Novel & HM & Base & Novel & HM & Base & Novel & HM \\
    \midrule
    CLIP       & 69.36 & 75.35 & 72.23 & 53.24 & 59.90 & 56.37 & 56.48 & 64.05 & 60.03 & 70.53 & 77.50 & 73.85 \\
    CoOp       & 80.60 & 65.89 & 72.51 & 79.44 & 41.18 & 54.24 & 92.19 & 54.74 & 68.69 & 84.69 & 56.05 & 67.46 \\
    CoCoOp     & 79.74 & 76.86 & 78.27 & 77.01 & 56.00 & 64.85 & 87.49 & 60.04 & 71.21 & 82.33 & 73.45 & 77.64 \\
    KgCoOp     & 80.29 & 76.53 & 78.36 & 77.55 & 54.99 & 64.35 & 85.64 & 64.34 & 73.48 & 82.89 & 76.67 & 79.66 \\
    MaPLe      & 80.82 & 78.70 & 79.75 & 80.36 & 59.18 & 68.16 & 94.07 & 73.23 & 82.35 & 83.00 & 78.66 & 80.77 \\
    PromptSRC  & \underline{82.67} & 78.47 & 80.52 & \underline{83.37} & 62.97 & 71.75 & 92.90 & 73.90 & 82.32 & 87.10 & 78.80 & 82.74 \\
    TCP        & 82.63 & 78.20 & 80.35 & 82.77 & 58.07 & 68.25 & 91.63 & 74.73 & 82.32 & \underline{87.13} & \underline{80.77} & \underline{83.83} \\
    MMA        & 82.27 & 78.57 & 80.38 & 83.20 & \underline{65.63} & \underline{73.38} & 85.46 & \textbf{82.34} & 83.87 & 86.23 & 80.03 & 82.20 \\
    CoPrompt   & 82.63 & \textbf{80.03} & \underline{81.31} & 83.13 & 64.73 & 72.79 & \underline{94.60} & 78.57 & \underline{85.84} & 86.90 & 79.57 & 83.07 \\
    \rowcolor{gray!20}\textbf{HiCroPL} & \textbf{83.23} & \underline{79.92} & \textbf{81.54} & \textbf{85.07} & \textbf{67.34} & \textbf{75.17} & \textbf{96.29} & \underline{80.36} & \textbf{87.61} & \textbf{87.95} & \textbf{80.91} & \textbf{84.28} \\
    \bottomrule
  \end{tabular}
}
  \caption{\textbf{Comparison with state-of-the-art methods on base-to-novel generalization.} The best results are bold-faced, with the second-best results underlined. The results demonstrate that the proposed HiCroPL achieves consistent improvement in domain adaptation and generalization.}
\label{tab:base2novel}
\vspace{-5pt}
\end{table*}

%-------------------------------------------------------------------------

%-------------------------------------------------------------------------

\section{Experiments}
\label{sec:experiments}
\subsection{Experiment Setup}
In line with previous works~\cite{CoOp, MaPLe}, we evaluate our approach on four benchmark settings. Due to page limitations, we provide a more detailed description of the dataset, training details, and LLM-generated templates in the Appendix.

\noindent\textbf{Base-to-novel Generalization.} Following the previous approaches~\cite{CoCoOp, MaPLe}, we split each dataset into base and novel classes. The model is trained on the base classes in a few-shot setting and evaluated on both the base and novel classes, with the harmonic mean (HM) reflecting their trade-off.

\noindent\textbf{Few-shot Learning.} We evaluate the model's ability to learn task-specific knowledge under limited supervision. The model's performance is assessed at various \textit{K}-shot settings, where \textit{K} = 1,2,4,8,16.

\noindent\textbf{Cross-dataset Evaluation.} In this setting, the model is trained on the source dataset ImageNet-1K with 16-shot training data and then directly evaluated on other datasets without any fine-tuning.

\noindent\textbf{Domain Generalization.} We evaluate the robustness of our approach on out-of-distribution datasets. The model is trained on ImageNet-1K and then directly evaluated on four variants of ImageNet datasets with different types of domain shifts.

%-------------------------------------------------------------------------

\begin{figure*}[t]
    \centering
    \includegraphics[width=\linewidth]{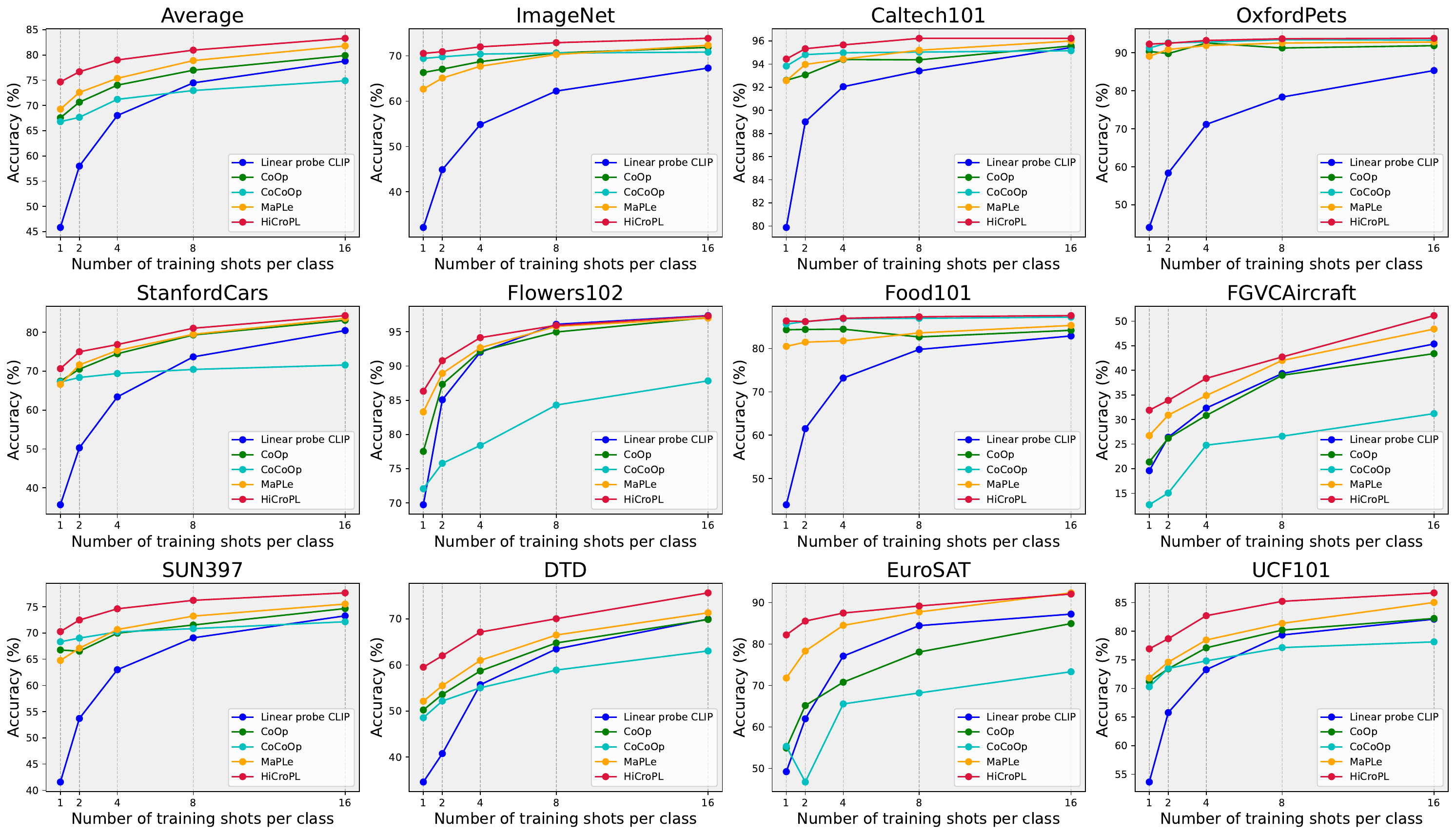}
    \vspace{-10pt}
    \caption{\textbf{HiCroPL performance comparison in few-shot image recognition setting.} HiCroPL demonstrates strong domain adaptability, indicating that the bidirectional knowledge flow effectively aligns representations between modalities.}  
    \label{fig:fewshot}
    \vspace{-5pt}
\end{figure*}

%-------------------------------------------------------------------------

\noindent\textbf{Implementation details.} Following previous works~\cite{CoOp, TCP, MaPLe}, all experiments adopt a ViT-B/16 CLIP backbone under 16-shot per-class training. For the base-to-novel generalization and few-shot learning tasks, we add prompts to all layers, setting their length to 16 and initializing them randomly with a normal distribution. The layer boundary \textit{k} is set to 6, meaning that in the first 6 layers, the prompts flow from text to image, while in the remaining 6 layers, the flow reverses from image to text. We use the LLM-generated category descriptions provided by CoPrompt\cite{CoPrompt} and set the consistency constraint $\lambda$ to 12. We train for 40 epochs with a batch size 128 on the large-scale ImageNet dataset. For the other ten datasets, we train for 50 epochs with a batch size 32. For the remaining two tasks, we train for only 5 epochs. The corresponding hyperparameters are fixed across all datasets in the same task. 

%-------------------------------------------------------------------------
\begin{table}[t]
  \centering
  \resizebox{\linewidth}{!}{
    \begin{tabular}{cc|ccccccccccc}
    \toprule
    \multirow{2}[4]{*}{} & Source & \multicolumn{11}{c}{Target} \\
\cmidrule{2-13}          & \rotatebox{90}{ImageNet} & \rotatebox{90}{Caltech101} & \rotatebox{90}{OxfordPets} & \rotatebox{90}{StanfordCars} & \rotatebox{90}{Flowers102} & \rotatebox{90}{Food101} & \rotatebox{90}{Aircraft} & \rotatebox{90}{SUN397} & \rotatebox{90}{DTD}   & \rotatebox{90}{EuroSAT} & \rotatebox{90}{UCF101} & \rotatebox{90}{Average} \\
    \midrule
    CoOp  & 71.51 & 93.70  & 89.14 & 64.51 & 68.71 & 85.30  & 18.47 & 64.15 & 41.92 & 46.39 & 66.55 & 63.88 \\
    CoCoOp & 71.02 & 94.43 & 90.14 & 65.32 & 71.88 & 86.06 & 22.94 & 67.36 & 45.73 & 45.37 & 68.21 & 65.74 \\
    MaPLe & 70.72 & \textbf{95.53} & \underline{90.49} & 65.57 & \underline{72.20}  & 86.20  & 24.74 & 67.01 & 46.49 & 48.06 & 68.69 & 66.30 \\
    MMA   & 71.00  & 93.80  & 90.30  & \textbf{66.13} & 72.07 & 86.12 & \underline{25.33} & \underline{68.17} & 46.57 & \underline{49.24} & 68.32 & 66.61 \\
    CoPrompt & 70.80  & \underline{94.50}  & \textbf{90.73} & 65.67 & \textbf{72.30}  & \underline{86.43} & 24.00    & 67.57 & \underline{47.07} & \textbf{51.90}  & \underline{69.73} & \underline{67.00} \\
    \rowcolor{gray!20}
    HiCroPL & 70.84 & 94.48 & 90.13 & \underline{65.68} & 72.03 & \textbf{86.46} & \textbf{26.58} & \textbf{68.78} & \textbf{53.19} & 49.19 & \textbf{70.31} & \textbf{67.68} \\
    \bottomrule
    \end{tabular}%
    }
    \caption{\textbf{Performance of HiCroPL on cross-dataset evaluation and its comparison to existing methods.} Overall, our method achieves the best average performance. Notably, on DTD, we observe a significant improvement, demonstrating the strong zero-shot transfer capability of our approach.
 }
  \label{tab:cross_dataset}%
\end{table}%

%-------------------------------------------------------------------------
\begin{table}[t]
\renewcommand{\arraystretch}{0.9}
  \centering
  \scriptsize
    \resizebox{\linewidth}{!}{
    \begin{tabular}{cc|ccccc}
    \toprule
          & Source & \multicolumn{5}{c}{Target} \\
\cmidrule{2-7}          & ImageNet & -V2 & -S & -A & -R & Avg. \\
    \midrule
    CoOp  & 71.51 & 64.2  & 47.99  & 49.71 & 75.21 & 59.28 \\
    CoCoOp & 71.02 & 64.07 & 48.75  & 50.63 & 76.18 & 59.91 \\
    MaPLe & 70.72 & 64.07 & 49.15  & \underline{50.90}  & 76.98 & 60.27 \\
    MMA   & 71.00  & \textbf{64.33} & 49.13  & \textbf{51.12} & \underline{77.32} & \textbf{60.48} \\
    CoPrompt & 70.80  & \underline{64.25} & \underline{49.43} & 50.50  & \textbf{77.51} & 60.42 \\
    \rowcolor{gray!20}
    HiCroPL & 71.22 & \textbf{64.33} & \textbf{49.47} & 50.79 & 77.15 & \underline{60.44} \\
    \bottomrule
    \end{tabular}%
    }
  \caption{\textbf{Performance on domain generalization.} Our method obtains the best performance on half of the datasets and achieves comparable average performance, showing good robustness to domain shifts.}
  \label{tab:ood}%
\end{table}%

%-------------------------------------------------------------------------

\subsection{Base-to-Novel Generalization}
Table~\ref{tab:base2novel} compares HiCroPL with 9 state-of-the-art methods (zero-shot CLIP~\cite{CLIP}, CoOp~\cite{CoOp}, CoCoOp~\cite{CoCoOp}, KgCoOp~\cite{KgCoOp}, MaPLe~\cite{MaPLe}, PromptSRC~\cite{PromptSRC}, TCP~\cite{TCP}, MMA~\cite{MMA}, CoPrompt~\cite{CoPrompt}) on the base-to-novel generalization task across 11 datasets. HiCroPL achieves consistent improvements of 1.89\% in base classes, 0.76\% in novel classes, and 1.28\% in harmonic mean over the previous best method, CoPrompt~\cite{CoPrompt}. Notably, this improvement does not compromise base class performance; instead, HiCroPL surpasses the second-best method, PromptSRC~\cite{PromptSRC}, by 1.63\% on base classes, highlighting its strong adaptation capability.

Compared to MaPLe~\cite{MaPLe}, the first method to explore inter-modal collaboration, HiCroPL improves by 3.61\%, 2.85\%, and 3.20\% on base, novel, and harmonic mean, respectively. This significant gain validates the effectiveness of our bidirectional knowledge flow over MaPLe’s one-way coupling, achieving deeper cross-modal alignment through semantic reciprocity.

%-------------------------------------------------------------------------

%-------------------------------------------------------------------------

\subsection{Few-shot Experiments} To evaluate in-domain generalization, we train HiCroPL under \textit{K}-shot supervision (\textit{K} = 1,2,4,8,16) and compare it with previous methods. As shown in Fig.~\ref{fig:fewshot}, HiCroPL consistently outperforms previous approaches achieving average gains of 5.40\%, 4.09\%, 3.64\%, 2.07\%, and 1.51\% across \textit{K} settings. Notably, HiCroPL demonstrates even more significant improvements in extreme low-shot scenarios (\textit{K} = 1, 2). This validates that HiCroPL’s bidirectional knowledge flow enables robust cross-modal alignment, even with minimal supervision. Detailed results for each dataset are provided in the Appendix.

\subsection{Cross-dataset Evaluation} We further assess HiCroPL's generalization by training on ImageNet and directly evaluating on 10 downstream datasets. As shown in Table~\ref{tab:cross_dataset}, HiCroPL achieves the best average performance and outperforms the second-best method CoPrompt on 6/10 datasets. The most significant gain of 6.12\% is observed on DTD~\cite{dtd}, demonstrating HiCroPL's exceptional zero-shot transfer capability.

\subsection{Domain Generalization} Table~\ref{tab:ood} shows HiCroPL's performance on out-of-distribution datasets. Our method achieves state-of-the-art results on half of the benchmarks while maintaining competitive average performance. This validates HiCroPL's ability to preserve transferable low-level representations, which is crucial for robust generalization under domain shifts.

%-------------------------------------------------------------------------
\begin{table}[t]
  \centering
    \resizebox{\linewidth}{!}{
    \begin{tabular}{ccccc}
    \toprule
    Mechanism & Knowledge Flow & Base  & Novel & HM \\
    \midrule
    \multirow{2}[2]{*}{Unidirectional} & I$\rightarrow$T  & 83.39 & 75.24  & 79.10  \\
          & T$\rightarrow$I  & 84.08 & 76.47  & 80.10 \\
    \cmidrule{1-5}    
    \multirow{2}[2]{*}{Bidirectional}  & I$\rightarrow$T $|$ T$\rightarrow$I & 85.44 & 76.23  & 80.58 \\
          & \cellcolor{gray!20}T$\rightarrow$I $|$ I$\rightarrow$T & \cellcolor{gray!20}\textbf{85.89} & \cellcolor{gray!20}\textbf{77.99} & \cellcolor{gray!20}\textbf{81.75} \\
    \bottomrule
    \end{tabular}%
    }
  \caption{Comparison of different knowledge flows configuration. T$\rightarrow$I indicates prompts flow from text to image, and ``$|$" means the layer boundary \textit{k} that controls when knowledge flow direction reverses.}
  \label{tab:knowledgeflow}%
\end{table}%

\begin{table}[t]
  \centering
   \resizebox{0.9\linewidth}{!}{
    \begin{tabular}{cccccc}
    \toprule
    Layer boundary  & k = 2 & k = 4 & k = 6 & k = 8 & k = 10 \\
    \toprule
    Base  & 85.76 & 85.82 & \textbf{85.89} & 85.49 & 85.19 \\
    Novel & 77.41 & 77.55 & \textbf{77.99} & 77.79 & 77.75 \\
    HM    & 81.37 & 81.48 & \textbf{81.75} & 81.46 & 81.30  \\
    \bottomrule
    \end{tabular}%
    }
    \caption{Ablation study of different layer boundary \textit{k}. Balanced knowledge interaction achieves the best performance.}
  \label{tab:k}%
\end{table}%

\begin{table}[t]
  \centering
    \begin{tabular}{cccc}
    \toprule
    Mapper design & Base  & Novel & HM \\
    \midrule
    Single-scale $|$ Single-scale  & 85.26 & 76.95 & 80.89  \\
    Single-scale $|$ Multi-scale  & 85.67 & 77.27 & 81.25  \\
    \rowcolor{gray!20}
    Multi-scale $|$ Multi-scale & \textbf{85.89} & \textbf{77.99} & \textbf{81.75}  \\
    \bottomrule
    \end{tabular}%
    \caption{Comparison of different mapper designs. Our multi-scale mapper works best.}
  \label{tab:mapperdesign}%
\end{table}%

\begin{table}[t]
  \centering
    \begin{tabular}{cccc}
    \toprule
    Compression Strategies & Base  & Novel & HM \\
    \midrule
    Equal weighting (averaging)   & 85.63 & 77.39 & 81.30 \\
    Multilayer perceptron (mlp)  & 85.06 & 77.68 & 81.20 \\
    \rowcolor{gray!20}
    LKP (ours)   & \textbf{85.89} & \textbf{77.99} & \textbf{81.75} \\
    \bottomrule
    \end{tabular}%
    \caption{Ablation on prompt compression techniques. Layer-specific knowledge proxy (LKP) provides better performance.}
  \label{tab:compress}%
    \vspace{-10pt}
\end{table}%
%-------------------------------------------------------------------------

\subsection{Ablative Analysis}
\noindent\textbf{Direction of knowledge flow.} In our proposed bidirectional knowledge flow mechanism, knowledge flows from text to image and then back from image to text. To evaluate the significance of this design, we compare the performance of different knowledge flow configurations. As shown in Table~\ref{tab:knowledgeflow}, bidirectional knowledge flow consistently outperforms unidirectional knowledge flow. This demonstrates that complete knowledge exchange is essential for effective cross-modal alignment. Furthermore, our approach achieves significant improvements over I$\rightarrow$T $|$ T$\rightarrow$I configuration. This is attributed to our design, which leverages the complementary strengths of different modalities at varying depths to iteratively refine each other's semantics.

\noindent\textbf{Layer partition analysis.} We analyze the layer boundary \textit{k}, which controls the reversal of knowledge flow. To evaluate its impact, we vary \textit{k} and measure performance across 11 datasets, the results are presented in Table~\ref{tab:k}. Our findings indicate that the optimal performance is achieved when \textit{k} = 6. In contrast, extreme values of \textit{k} (\eg, \textit{k} = 2 or \textit{k} = 10) lead to a degradation in accuracy for novel and base classes by 0.58\% and 0.70\%, respectively. This highlights the importance of balanced interactions between modalities. 

\noindent\textbf{Effect of multi-scale mapping.} The hierarchical knowledge mapping ensures that the prompts at each layer can absorb knowledge from multiple scales, enabling the final decision-making representations to incorporate rich intermediate features. We conduct an ablation study by replacing our component with a single-scale projection, similar to MaPLe~\cite{MaPLe}. The results in Table~\ref{tab:mapperdesign} demonstrate that multi-scale knowledge mapping improves the model's generalization ability.

\noindent\textbf{Effect of LKP.} We ablate on the choice of prompt compression techniques. Specifically, we consider two alternatives to LKP: assigning equal weights to all prompts and using a 2-layer MLP for fusion, with the results presented in Table~\ref{tab:compress}. Our LKP achieves the best performance, as it dynamically selects the importance of layer-specific knowledge. Compared to treating each prompt equally, LKP better preserves key semantics while filtering out noise.

\noindent\textbf{Training and inference cost analysis.} In Table~\ref{tab:efficiency_analysis}, we show the compute cost analysis of our approach. During training, our approach requires 7.9\% more training time than MaPLe due to the need for generating supervision features from the pretrained model. However, our inference GFLOPs are only 0.014$\times$ higher than MaPLe, while achieving a remarkable 3.2\% absolute gain. Additionally, our LKP compresses inter-layer prompts, further enhancing efficiency.

\begin{figure}[t]
    \centering
    \includegraphics[width=\linewidth]{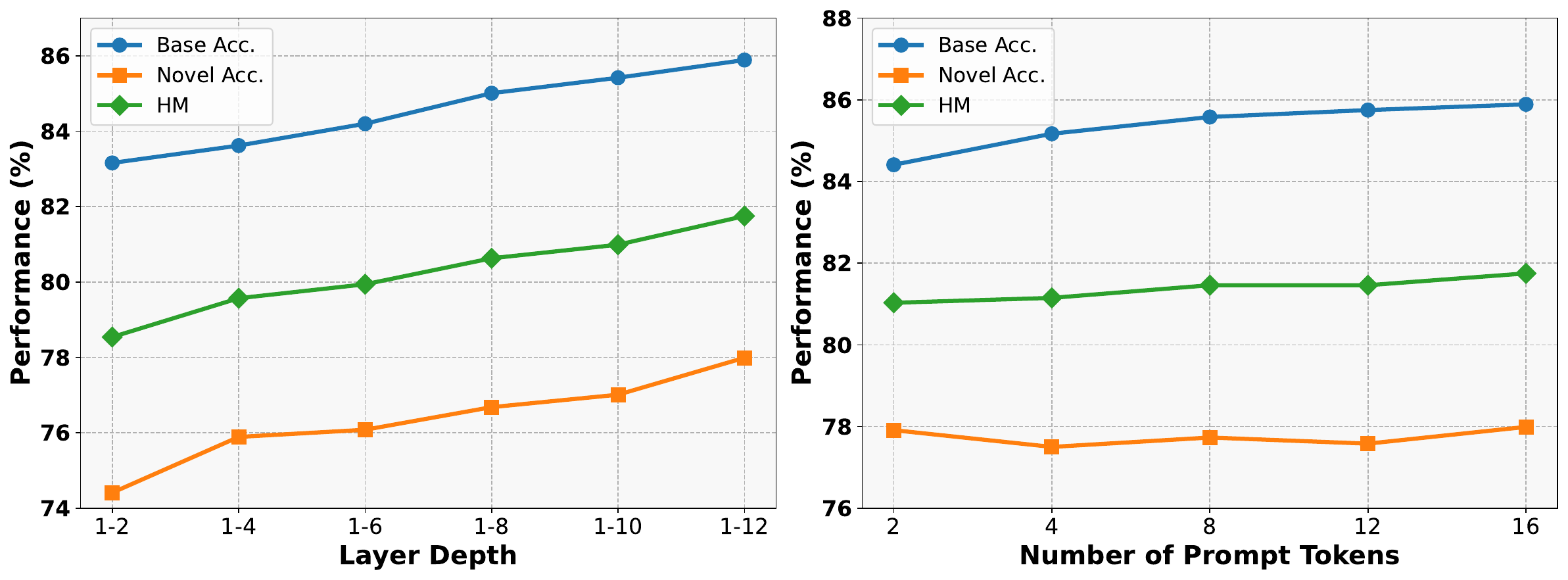}
    \vspace{-20pt}
    \caption{Ablation on prompt depth (\textit{\textbf{left}}) and prompt length (\textit{\textbf{right}}) in HiCroPL.}
    \label{fig:parameter}
    \vspace{-10pt}
\end{figure}

\noindent\textbf{Prompt Depth.} Fig.~\ref{fig:parameter} (left) shows the effect of prompt depth for HiCroPL. Overall, the performance improves as prompt depth increases, HiCroPL achieves maximum performance at a depth of 12.

\noindent\textbf{Prompt Length.} In Fig.~\ref{fig:parameter} (right), we illustrate the effect of prompt length for our proposed method. As the prompt length increases, the performance on base classes rises relatively significantly, while the novel classes have remained relatively stable.

%-------------------------------------------------------------------------

\begin{table}[t]
  \centering
    \begin{tabular}{lcccc}
    \toprule
    Method  & GFLOPs (test) & Train time (min) & HM \\
    \midrule
    MaPLe      &  108.28     & 20.44 &  78.55 \\
    HiCroPL*       &  109.95     & 23.53 &  81.63 \\
   \rowcolor{gray!20}  
   HiCroPL       &  109.81     & 22.22 &  \textbf{81.75} \\
    \bottomrule
    \end{tabular}%
      \caption{Efficiency analysis of compute cost. Training time is calculated for 40 epochs on a single A100 GPU on the ImageNet dataset. HiCroPL* denotes the implementation without LKP.}
  \label{tab:efficiency_analysis}%
\end{table}%

%-------------------------------------------------------------------------

\section{Conclusion} 
Prompt learning has been shown to effectively adapt VLMs like CLIP to downstream tasks. However, two key bottlenecks limit the generalization ability of existing prompt learning methods: (a) modality isolation and (b) hierarchical semantic decay. In this work, we introduce HiCroPL, which addresses both challenges and achieves better generalization. Our results demonstrate that enabling bidirectional interaction between modalities during fine-tuning is crucial for improving cross-modal alignment and refining semantics between modalities. Additionally, we propose a hierarchical knowledge mapper that allows different scale representations to merge during the mapping process, ensuring that transferable low-level representations in intermediate layers contribute to task decisions, thereby enhancing the model’s generalization ability. Extensive evaluations across four different tasks show that HiCroPL outperforms existing state-of-the-art methods across zero-shot learning, few-shot learning, cross-dataset, and domain generalization tasks, achieving significant improvements.

\noindent\textbf{Acknowledgement.} We would like to thank all reviewers for their constructive comments and suggestions. This work was supported by the Natural Science Foundation of China (No.62172166) and the Guangdong Basic and Applied Basic Research Foundation (No.2022A1515011380).

%-------------------------------------------------------------------------
{
    \small
    \bibliographystyle{ieeenat_fullname}
    \bibliography{main}
}


\clearpage
\setcounter{page}{1}
\maketitlesupplementary
\appendix

The following sections contain supplemental information and encompass the formulation of the Hierarchical Knowledge Mapper in Sec.~\ref{appendix:formal_description}, more implementation details in Sec.~\ref{appendix:additional_implementation_details}, and a thorough ablative analysis of HiCroPL in Sec~\ref{appendix:additional_Experiments}. 
\section{Formal Description of Hierarchical Knowledge Mapper}
\label{appendix:formal_description}
The hierarchical knowledge mapper projects multi-scale knowledge into a single prompt of another modality, which allows the prompt to adaptively absorb cross-modal information from multiple scales. Taking text-to-image mapping as an example, formally, let $P_v = \{p^{l,1}_v, p^{l,2}_v,...,p^{l,m}_v\} \in \mathbb{R}^{k \times m \times d_v}$ denote visual prompts and $\tilde{P_p} = \{\tilde{p^{1}_p},\tilde{p^{2}_p},...,\tilde{p^{k}_p}\}$ represent refined textual proxy tokens. The cross-modal mapping is computed as:
\begin{equation}
\begin{aligned}
\mathbf{Q} &= P_v \mathbf{W}_q, \quad \mathbf{W}_q \in \mathbb{R}^{d_v \times d_v}, \\
\mathbf{K} &= P_p \mathbf{W}_k, \quad \mathbf{W}_k \in \mathbb{R}^{d_t \times d_v}, \\
\mathbf{V} &= P_p \mathbf{W}_v, \quad \mathbf{W}_v \in \mathbb{R}^{d_t \times d_v},
\end{aligned}
\end{equation}
where $\mathbf{W}_q, \mathbf{W}_k, \mathbf{W}_v$ are learnable projection matrices. The scaled dot-product attention computes cross-modal interaction:
\begin{equation}
\text{Attention}(\mathbf{Q},\mathbf{K},\mathbf{V}) = \text{Softmax}\left(\frac{\mathbf{Q}\mathbf{K}^\top}{\sqrt{d_v}}\right)\mathbf{V}.
\end{equation}
Following the standard transformer architecture, we employ layer normalization and residual connections:
\begin{equation}
\begin{aligned}
\mathbf{Q}' &= \mathbf{Q} + \text{Attention}(\text{LN}(\mathbf{Q}),\text{LN}(\mathbf{K}),\text{LN}(\mathbf{V})), \\
P_v &= \mathbf{Q}' + \text{FFN}(\text{LN}(\mathbf{Q}')),
\end{aligned}
\end{equation}
where FFN denotes the feed-forward network with GELU activation:
\begin{equation}
\text{FFN}(\mathbf{x}) = \mathbf{W}_2 \cdot \text{GELU}(\mathbf{W}_1 \mathbf{x} + \mathbf{b}_1) + \mathbf{b}_2,
\end{equation}
where $\mathbf{W}_1$, $\mathbf{W}_2$, $\mathbf{b}_1$, and $\mathbf{b}_2$ are learnable parameters.

\section{Additional Implementation Details}
\label{appendix:additional_implementation_details}
\noindent \textbf{Additional Training details.} We train HiCroPL for 5 epochs for cross-dataset evaluation and domain generalization settings. The text feature dimension $d_t$ = 512 and the image feature dimension $d_v$ = 768.  We fix the learning rate at 0.0025, and optimization is performed using the Adam optimizer with a momentum of 0.9 and weight decay of 0.0005. The corresponding hyperparameters are fixed across all datasets in the same task. All experiments are conducted on a single NVIDIA A100 GPU.

\begin{table}[t]
  \centering
    \resizebox{\linewidth}{!}{
    \begin{tabular}{c|cc|c|c|c|c|c|c|c}
    \toprule
    \multicolumn{1}{c}{Dataset} & \multicolumn{1}{c}{Class name} & \multicolumn{8}{c}{LLM-generated descriptions.} \\
    \midrule
    \multirow{6}[6]{*}{SUN397} & \multirow{2}[2]{*}{airplane cabin} & \multicolumn{8}{c}{\multirow{2}[2]{*}{\thead{The cabin of an airplane typically has rows \\ of seats on either side of a central aisle.}}} \\
          &       & \multicolumn{8}{c}{} \\
\cmidrule{2-10}          & \multirow{2}[2]{*}{bookstore} & \multicolumn{8}{c}{\multirow{2}[2]{*}{\thead{A bookstore has shelves full of books and usually \\ has a desk where you can pay for your books.}}} \\
          &       & \multicolumn{8}{c}{} \\
\cmidrule{2-10}          & \multirow{2}[2]{*}{campus} & \multicolumn{8}{c}{\multirow{2}[2]{*}{\thead{A campus looks like a collection \\ of buildings that are close together.}}} \\
          &       & \multicolumn{8}{c}{} \\
    \bottomrule
    \end{tabular}%
    }
    \caption{Example of descriptive text generated by LLM.}
  \label{tab:LLMdata}%
\end{table}%

\begin{table}[t]
  \centering
  \resizebox{\linewidth}{!}{
    \begin{tabular}{ccccc}
    \toprule
    \multicolumn{1}{c}{Datasets} & \multicolumn{1}{c}{Classes} & Training Size & Validation Size & Testing Size \\
    \midrule
    ImageNet~\cite{imagenet}  & 1,000 & 1,281,167 & N/A   & 50,000 \\
    Caltech101~\cite{caltech101} & 100   & 4,128 & 1,649 & 2,465 \\
    EuroSAT~\cite{eurosat}  & 10    & 13,500 & 5,400 & 8,100 \\
    SUN397~\cite{sun397}  & 397   & 15,880 & 3,970 & 19,850 \\
    DTD~\cite{dtd}   & 47    & 2,820 & 1,128 & 1,692 \\
    UCF101~\cite{ucf101}  & 101   & 7,639 & 1,808 & 3,783 \\
    FGVCAircraft~\cite{fgvcaircraft} & 100   & 3,334 & 3,333 & 3,333 \\
    OxfordPets~\cite{oxfordpets} & 37    & 2,944 & 736   & 3,669 \\
    StanfordCars~\cite{stanfordcars} & 196   & 6,509 & 1,635 & 8,041 \\
    Flowers102~\cite{flowers102} & 102   & 4,093 & 1,633 & 2,463 \\
    Food101~\cite{food101} & 101   & 50,500 & 20,200 & 30,300 \\
    \midrule
    ImageNet-V2~\cite{imagenet-v2} & 1000  & N/A   & N/A   & 10000 \\
    ImageNet-Sketch~\cite{imagenet-sketch} & 1000  & N/A   & N/A   & 50889 \\
    ImageNet-A~\cite{imagenet-a} & 200   & N/A   & N/A   & 7500 \\
    ImageNet-R~\cite{imagenet-r} & 200   & N/A   & N/A   & 30000 \\
    \bottomrule
    \end{tabular}%
    }
  \caption{Detailed statistics of the datasets.}
  \label{tab:datasets}%
\end{table}%

\noindent \textbf{LLM-generated category descriptions.} We employ large language model (LLM) to generate detailed descriptions for each category, providing diverse frozen text features. For each category, we utilize GPT-3~\cite{brown2020language} to generate descriptive sentences. For simplicity, we adopt the publicly available CoPrompt~\cite{CoPrompt} data. However, unlike CoPrompt, we average the embeddings of all descriptions for each category to obtain the final category embedding, rather than dynamically selecting a single sentence as the category representation. Table~\ref{tab:LLMdata} presents a sample of the LLM-generated category descriptions.

\noindent \textbf{Datasets.} We evaluate the performance of our method on 15 recognition datasets. For base-to-novel generalization and cross-dataset evaluation tasks, we evaluate our method on 11 image datasets covering various recognition tasks. These include ImageNet~\cite{imagenet} and Caltech101~\cite{caltech101} for general object recognition. Five fine-grained classification datasets, OxfordPets~\cite{oxfordpets}, StanfordCars~\cite{stanfordcars}, Flowers102~\cite{flowers102}, Food101~\cite{food101}, and FGVCAircraft~\cite{fgvcaircraft}. SUN397~\cite{sun397} is used for scene recognition, UCF101~\cite{ucf101} for action recognition, DTD~\cite{dtd} for texture classification, and EuroSat~\cite{eurosat} for satellite image classification. For the domain generalization task, ImageNet~\cite{imagenet} is used as the source domain dataset for training the model, and its variants ImageNet-A~\cite{imagenet-a}, ImageNet-R~\cite{imagenet-r}, ImageNet-Sketch~\cite{imagenet-sketch} and ImageNet-V2~\cite{imagenet-v2} are used for out-of-distribution dataset evaluation. The detailed statistics of the 11 datasets, as well as the four variants of ImageNet~\cite{imagenet}, are shown in Table~\ref{tab:datasets}.

%-------------------------------------------------------------------------

\begin{table}[t]
  \centering
    \begin{tabular}{ccccc}
    \toprule
    BKF   & $\mathcal{L}_{cons}$  & Base  & Novel & HM \\
    \midrule
          &       & 82.15 & 74.07 & 77.90 \\
          &   $\surd$    & 82.09 & 76.02 & 78.94 \\
      $\surd$    &       & \textbf{85.96}  & 74.65  & 79.91  \\
      \rowcolor{gray!20}
      $\surd$    &  $\surd$    & 85.89 & \textbf{77.99} & \textbf{81.75} \\
    \bottomrule
    \end{tabular}%
    \caption{Ablation experiments on the components of HiCroPL. BKF refers to the Bidirectional Knowledge Flow mechanism.}
  \label{tab:ablation_conp}%
\end{table}%

\begin{table}[t]
  \centering
    \begin{tabular}{cccc}
    \toprule 
    \multicolumn{1}{l}{Frozen prompts choice} & Base  & Novel & HM \\
    \midrule
    a photo of a \{\} & 84.92 & 75.99 & 80.21 \\
    frozen diverse prompts & 85.14 & 75.23 & 79.88 \\
    LLM (a sentence) & 85.33 & 76.41 & 80.63 \\
    \rowcolor{gray!20}
    LLM(ensemble) & \textbf{85.89} & \textbf{77.99} & \textbf{81.75}  \\
    \bottomrule
    \end{tabular}%
    \caption{Ablation on frozen prompt choices. }
  \label{tab:frozen_prompt}%
\end{table}%

%-------------------------------------------------------------------------

\section{Additional Experiments}
\label{appendix:additional_Experiments}
\noindent \textbf{Effect of consistency regularization.} Table~\ref{tab:ablation_conp} provides ablation experiments on the components of HiCroPL. The bidirectional knowledge flow mechanism significantly boosts base class performance and achieves the best overall results. Additionally, by leveraging intermediate-layer features, it also improves performance on novel classes. While using the regularization term alone enhances generalization to novel classes, it does not provide gains on base classes. Ultimately, the combination of both components in HiCroPL achieves the best performance.

\noindent \textbf{Effect of frozen prompts.} Since different frozen prompts provide distinct knowledge to constrain prompt learning, we evaluate the effectiveness of various hand-crafted prompts. Specifically, we compare the fixed prompt ``a photo of a \{\}" used in KgCoOp~\cite{KgCoOp}, the diverse textual descriptions in PromptSRC~\cite{PromptSRC}, the randomly sampled LLM prompts in CoPrompt~\cite{CoPrompt}, and the averaged LLM prompts in our HiCroPL. The results are shown in Table~\ref{tab:frozen_prompt}. Compared to the dynamically generated individual sentences in CoPrompt, ensemble LLM-generated prompts provide richer textual features, thereby improving performance. However, the diverse textual descriptions used in PromptSRC are based on the text templates provided by CLIP for ImageNet, which may lead to inaccurate descriptions when applied to other datasets, resulting in performance degradation.

\noindent \textbf{Influence of different consistency criteria.} We evaluate the impact of different consistency criteria on constraints in Table~\ref{tab:criteria}. The results show that using cosine similarity as the consistency criterion provides the best performance, followed by L1, while using MSE severely degrades the performance.

%-------------------------------------------------------------------------

\begin{table}[t]
  \centering
    \begin{tabular}{cccc}
    \toprule
    Criterion & Base  & Novel & HM \\
    \midrule
    MSE   & 85.11 & 74.39 & 79.39 \\
    L1    & 85.79 & 77.2  & 81.27 \\
    \rowcolor{gray!20}
    Cosine & \textbf{85.89} & \textbf{77.99} & \textbf{81.75}  \\
    \bottomrule
    \end{tabular}%
    \caption{Comparison of different distillation consistency criteria. Cosine similarity works best.}
  \label{tab:criteria}%
\end{table}%

%-------------------------------------------------------------------------

\noindent \textbf{Few-shot experiments.} We evaluate the adaptability of HiCroPL through few-shot experiments. Table~\ref{tab:fewshot} provides detailed per-dataset results for various methods under the few-shot setting. Compared to previous methods, HiCroPL achieves consistent improvements.

\definecolor{GrayColor}{gray}{0.85}
\begin{table*}[t]
\renewcommand{\arraystretch}{0.9}
\scriptsize
  \centering
    \begin{tabular}{>{\centering}p{8em}p{8em}|>{\centering}p{5em}>{\centering}p{5em}>{\centering}p{5em}>{\centering}p{5em}c}
    
    \toprule
    {\textbf{Dataset}} & {\textbf{Method}} & {{\textbf{1 shot}}} & {{\textbf{2 shots}}} & {{\textbf{4 shots}}} & {{\textbf{8 shots}}} & {{\textbf{16 shots}}} \\
    \midrule
    \multicolumn{1}{c}{\multirow{6}[2]{*}{{Average}}} & {Linear probe CLIP} & {45.83} & {57.98} & {68.01} & {74.47} & {78.79} \\
          & {CoOp} & {67.56} & {70.65} & {74.02} & {76.98} & {79.89} \\
          & {CoCoOp} & {66.79} & {67.65} & {71.21} & {72.96} & {{74.90}} \\
          & {MaPLe} & 69.27 & 72.58 & 75.37 & 78.89 & 81.79 \\
          & {PromptSRC} & {72.32} & {75.29} & {78.35} & {80.69} & {82.87} \\
          \rowcolor{GrayColor} & {HiCroPL} & \textbf{74.67} & \textbf{76.67} & \textbf{79.01} & \textbf{80.96} & \textbf{83.30} \\
    \midrule
    \multicolumn{1}{c}{\multirow{6}[2]{*}{{ImageNet}}} & {Linear probe CLIP} & {32.13} & {44.88} & {54.85} & {62.23} & {67.31} \\
          & {CoOp} & {66.33} & {67.07} & {68.73} & {70.63} & {71.87} \\
          & {CoCoOp} & {69.43} & {69.78} & {70.39} & {70.63} & {70.83} \\
          & {MaPLe} & {62.67} & {{65.10}} & {{67.70}} & {{70.30}} & {72.33} \\
          & {PromptSRC} & {68.13} & {69.77} & {71.07} & {72.33} & {73.17} \\
          \rowcolor{GrayColor} & {HiCroPL} & \textbf{70.54} & \textbf{70.92} & \textbf{71.99} & \textbf{72.91} & \textbf{73.87} \\
    \midrule
    \multicolumn{1}{c}{\multirow{6}[2]{*}{{Caltech101}}} & {Linear probe CLIP} & {79.88} & {89.01} & {92.05} & {93.41} & {95.43} \\
          & {CoOp} & {{92.60}} & {93.07} & {94.4} & {94.37} & {95.57} \\
          & {CoCoOp} & {93.83} & {94.82} & {94.98} & {95.04} & {95.16} \\
          & {MaPLe} & {92.57} & {93.97} & {94.43} & {{95.20}} & {{96.00}} \\
          & {PromptSRC} & {93.83} & {94.53} & {95.27} & {95.67} & {96.07} \\
          \rowcolor{GrayColor} & {HiCroPL} & \textbf{94.44} & \textbf{95.33} & \textbf{95.66} & \textbf{96.23} & \textbf{96.23} \\
    \midrule
    \multicolumn{1}{c}{\multirow{6}[2]{*}{{OxfordPets}}} & {Linear probe CLIP} & {44.06} & {58.37} & {71.17} & {78.36} & {85.34} \\
          & {CoOp} & {90.37} & {89.80} & {92.57} & {91.27} & {91.87} \\
          & {CoCoOp} & {91.27} & {92.64} & {92.81} & {93.45} & {93.34} \\
          & {MaPLe} & {{89.10}} & {90.87} & {{91.90}} & {92.57} & {92.83} \\
          & {PromptSRC} & {{92.00}} & {\textbf{92.50}} & \textbf{93.43} & {{93.50}} & {93.67} \\
          \rowcolor{GrayColor} & {HiCroPL} & \textbf{92.29} & \textbf{92.50} & 93.24 & \textbf{93.70} & \textbf{93.81} \\
    \midrule
    \multicolumn{1}{c}{\multirow{6}[2]{*}{{StanfordCars}}} & {Linear probe CLIP} & {35.66} & {50.28} & {63.38} & {73.67} & {80.44} \\
          & {CoOp} & {67.43} & {{70.50}} & {74.47} & {{79.30}} & {83.07} \\
          & {CoCoOp} & {67.22} & {68.37} & {69.39} & {70.44} & {71.57} \\
          & {MaPLe} & {{66.60}} & {{71.60}} & {{75.30}} & {79.47} & {83.57} \\
          & {PromptSRC} & {{69.40}} & {{73.40}} & \textbf{77.13} & {80.97} & {83.83} \\
          \rowcolor{GrayColor} & {HiCroPL} & \textbf{70.64} & \textbf{74.98} & 76.84 & \textbf{81.03} & \textbf{84.28} \\
    \midrule
    \multicolumn{1}{c}{\multirow{6}[2]{*}{{Flowers102}}} & {Linear probe CLIP} & {69.74} & {85.07} & {92.02} & {{96.10}} & {97.37} \\
          & {CoOp} & {77.53} & {87.33} & {92.17} & {94.97} & {97.07} \\
          & {CoCoOp} & {72.08} & {75.79} & {{78.40}} & {{84.30}} & {87.84} \\
          & {MaPLe} & {{83.30}} & {88.93} & {92.67} & {{95.80}} & {{97.00}} \\
          & {PromptSRC} & {85.93} & \textbf{91.17} & {93.87} & \textbf{96.27} & {\textbf{97.60}} \\
          \rowcolor{GrayColor} & {HiCroPL} & \textbf{86.32} & 90.78 & \textbf{94.15} & 95.94 & 97.32 \\
    \midrule
    \multicolumn{1}{c}{\multirow{6}[2]{*}{{Food101}}} & {Linear probe CLIP} & {43.96} & {61.51} & {73.19} & {79.79} & {{82.90}} \\
          & {CoOp} & {84.33} & {{84.40}} & {84.47} & {82.67} & {{84.20}} \\
          & {CoCoOp} & {85.65} & \textbf{86.22} & {86.88} & {86.97} & {87.25} \\
          & {MaPLe} & {{80.50}} & {81.47} & {81.77} & {{83.60}} & {85.33} \\
          & {PromptSRC} & {84.87} & {{85.70}} & {86.17} & {{86.90}} & {{87.50}} \\
          \rowcolor{GrayColor} & {HiCroPL} & \textbf{86.37} & 86.21 & \textbf{86.98} & \textbf{87.33} & \textbf{87.6} \\
    \midrule
    \multicolumn{1}{c}{\multirow{6}[2]{*}{{FGVCAircraft}}} & {Linear probe CLIP} & {19.61} & {26.41} & {32.33} & {39.35} & {45.36} \\
          & {CoOp} & {21.37} & {{26.20}} & {30.83} & {{39.00}} & {{43.40}} \\
          & {CoCoOp} & {12.68} & {15.06} & {24.79} & {26.61} & {31.21} \\
          & {MaPLe} & {26.73} & {{30.90}} & {34.87} & {{42.00}} & {{48.40}} \\
          & {PromptSRC} & {27.67} & {{31.70}} & {37.47} & \textbf{43.27} & {50.83} \\
          \rowcolor{GrayColor} & {HiCroPL} & \textbf{31.89} & \textbf{33.90} & \textbf{38.37} & 42.72 & \textbf{51.13} \\
    \midrule
    \multicolumn{1}{c}{\multirow{6}[2]{*}{{SUN397}}} & {Linear probe CLIP} & {41.58} & {{53.70}} & {{63.00}} & {69.08} & {73.28} \\
          & {CoOp} & {66.77} & {66.53} & {69.97} & {71.53} & {74.67} \\
          & {CoCoOp} & {68.33} & {69.03} & {70.21} & {70.84} & {72.15} \\
          & {MaPLe} & {64.77} & {{67.10}} & {70.67} & {73.23} & {75.53} \\
          & {PromptSRC} & {69.67} & {{71.60}} & {{74.00}} & {75.73} & {77.23} \\
          \rowcolor{GrayColor} & {HiCroPL} & \textbf{70.27} & \textbf{72.48} & \textbf{74.62} & \textbf{76.24} & \textbf{77.66} \\
    \midrule
    \multicolumn{1}{c}{\multirow{6}[2]{*}{{DTD}}} & {Linear probe CLIP} & {34.59} & {40.76} & {55.71} & {63.46} & {69.96} \\
          & {CoOp} & {50.23} & {{53.60}} & {{58.70}} & {64.77} & {69.87} \\
          & {CoCoOp} & {48.54} & {52.17} & {55.04} & {58.89} & {63.04} \\
          & {MaPLe} & {52.13} & {55.5} & {{61.00}} & {{66.50}} & {71.33} \\
          & {PromptSRC} & {56.23} & {59.97} & {65.53} & {69.87} & {72.73} \\
          \rowcolor{GrayColor} & {HiCroPL} & \textbf{59.52} & \textbf{62.00} & \textbf{67.14} & \textbf{70.04} & \textbf{75.65} \\
    \midrule
    \multicolumn{1}{c}{\multirow{6}[2]{*}{{EuroSAT}}} & {Linear probe CLIP} & {49.23} & {61.98} & {77.09} & {84.43} & {87.21} \\
          & {CoOp} & {54.93} & {65.17} & {{70.80}} & {78.07} & {84.93} \\
          & {CoCoOp} & {55.33} & {46.74} & {65.56} & {68.21} & {73.32} \\
          & {MaPLe} & {{71.80}} & {{78.30}} & {{84.50}} & {87.73} & {92.33} \\
          & {PromptSRC} & {73.13} & {79.37} & {{86.90}} & {{88.80}} & \textbf{92.43} \\
          \rowcolor{GrayColor} & {HiCroPL} & \textbf{82.2} & \textbf{85.53} & \textbf{87.47} & \textbf{89.17} & 92.05 \\
    \midrule
    \multicolumn{1}{c}{\multirow{6}[2]{*}{{UCF101}}} & {Linear probe CLIP} & {53.66} & {65.78} & {73.28} & {79.34} & {82.11} \\
          & {CoOp} & {71.23} & {73.43} & {{77.10}} & {{80.20}} & {82.23} \\
          & {CoCoOp} & {{70.30}} & {73.51} & {74.82} & {77.14} & {78.14} \\
          & {MaPLe} & {71.83} & {{74.60}} & {78.47} & {81.37} & {85.03} \\
          & {PromptSRC} & {{74.80}} & {{78.50}} & {81.57} & {{84.30}} & {86.47} \\
          \rowcolor{GrayColor} & {HiCroPL} & \textbf{76.92} & \textbf{78.69} & \textbf{82.71} & \textbf{85.22} & \textbf{86.70} \\
    \bottomrule
    \end{tabular}%
     \caption{Comparison of HiCroPL performance with various methods for each dataset in few-shot setting.}
  \label{tab:fewshot}%
\end{table*}%

\end{document}